\documentclass[times,twocolumn,final]{elsarticle}

\usepackage{comment}
\usepackage{medima}
\usepackage{framed,multirow}
\usepackage{amsmath}
 \usepackage{epstopdf}
 \usepackage{epsfig}
 \usepackage{bm}
\usepackage{amssymb}
\usepackage{latexsym}
\usepackage{multirow}
\usepackage{url}
\usepackage{xcolor}
 \usepackage{verbatim}
\usepackage{hyperref}
\usepackage{algorithmic}
\usepackage{graphicx}
\usepackage{textcomp}
\usepackage{amsmath,amssymb,amsfonts}

\journal{Medical Image Analysis}

\begin{document}

\verso{C. Zhan         \textit{et~al.}}

\begin{frontmatter}

\title{UnICLAM:Contrastive Representation Learning with Adversarial Masking for Unified and Interpretable Medical Vision Question Answering}%
\fntext[fn1]{These authors contributed equally and share the first authorship.}
\author[1]{Chenlu \snm{Zhan}\fnref{fn1}}
\ead{chenlu.22@intl.zju.edu}
\author[1]{Peng \snm{Peng}\fnref{fn1}}
\ead{pengpeng@intl.zju.edu.cn}

\author[2]{Hongsen \snm{Wang}}
\ead{whsin301@163.com}
\author[1]{Gaoang \snm{Wang}}
\ead{gaoangwang@intl.zju.edu.cn}

\author[1]{Yu Lin}
\ead{yulin@intl.zju.edu.cn}

\author[2]{Tao \snm{Chen}\corref{cor1}}
\ead{chentao301@126.com}
\author[1]{Hongwei \snm{Wang}\corref{cor1}}
\ead{hongweiwang@intl.zju.edu.cn}
\cortext[cor1]{Corresponding authors.}


\address[1]{Zhejiang University and the University of Illinois Urbana–Champaign Institute, Haining, 314400,
China.}
\address[2]{Sixth Medical Center of Chinese PLA General Hospital, Beijing, 100037, China.}


\begin{abstract}
Medical Visual Question Answering (Medical-VQA) aims to assist doctors' decision-making in answering  clinical questions regarding radiology images.
Nevertheless, current Medical-VQA models learn cross-modal representations through residing vision and texture encoders in dual separate spaces, which inevitably lead to indirect semantic alignment. 
In this paper, we propose UnICLAM, a Unified and Interpretable Medical-VQA model through  Contrastive Representation Learning with Adversarial Masking. To achieve the learning of an aligned image-text representation for UnICLAM, we first establish a unified dual-stream pre-training structure with the gradually soft-parameter sharing strategy. Specifically, the proposed strategy learns a constraint for the vision and texture encoders to be close as much as possible in the same space, which is gradually loosened as the number of layers increases.
For grasping the unified semantic representation, we extend the adversarial masking data augmentation to the contrastive representation learning of vision and text in a unified manner. While the encoder training minimizes the distance between the original and masking samples, the adversarial masking module keeps adversarial learning to conversely maximize the distance. Furthermore, we also intuitively take a further exploration of the unified adversarial masking augmentation method, which improves the potential \emph{ante-hoc} interpretability with remarkable performance and efficiency.
Experimental results on VQA-RAD and SLAKE public benchmarks demonstrate that UnICLAM outperforms existing $11$ state-of-the-art Medical-VQA methods.  More importantly, we make an additional discussion about the performance of UnICLAM in diagnosing heart failure, verifying that UnICLAM exhibits superior few-shot adaption performance in practical disease diagnosis. The codes and models will be released upon the acceptance of the paper.
\end{abstract}


\end{frontmatter}


\section{Introduction}
\label{sec1}
{M}{edical} Vision Question Answering (Medical-VQA) is a branch of general Vision Question Answering (VQA) that aims to product a correct answer with a given medical image and its corresponding question. The Medical-VQA requires building a cross-modal model that aligns the semantics from vision and texture representations, which is still in its infancy.

\begin{figure}[htbp]
\centering
\includegraphics[width=1\linewidth]{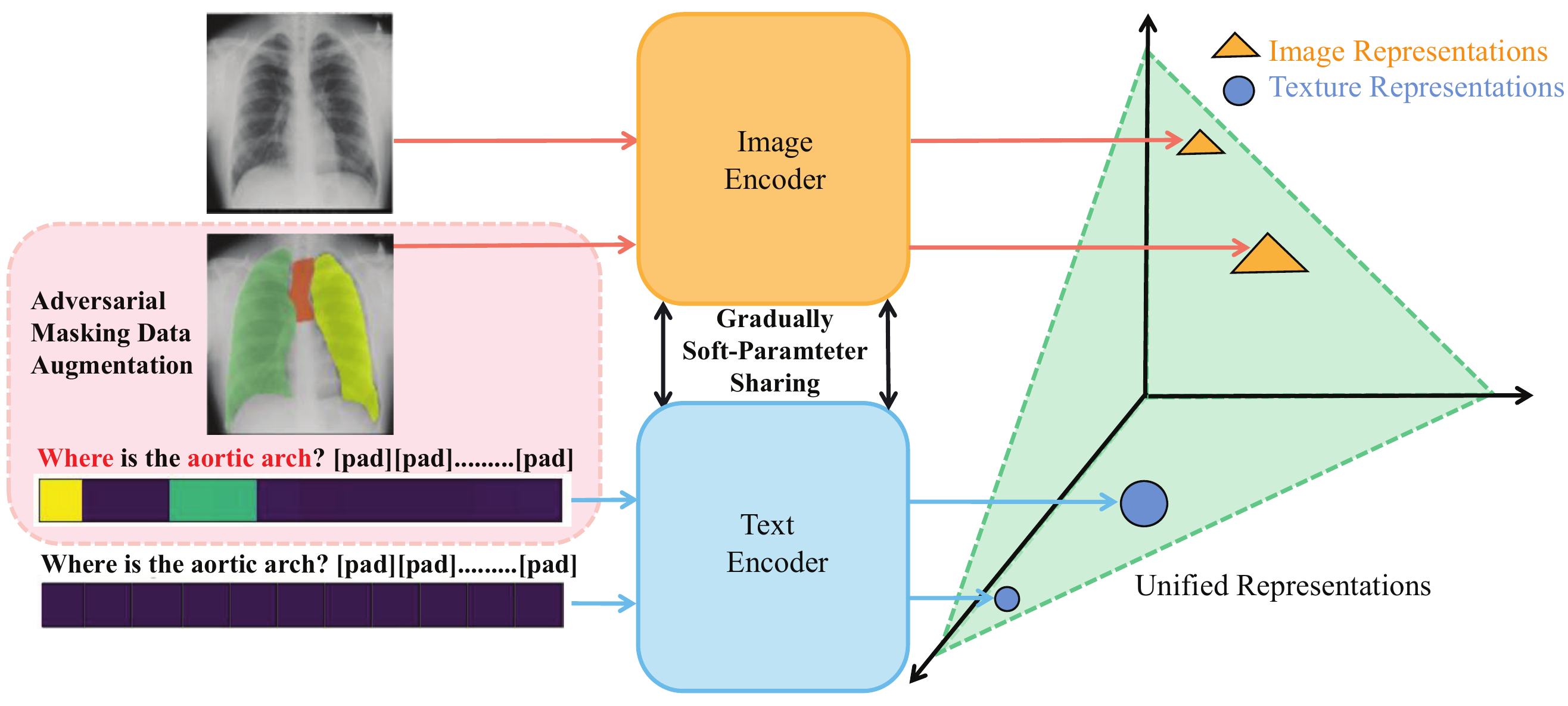}
\caption{Demonstration of the proposed unified and interpretable structure. Generally, the image encoder and texture encoder are constrained in the same space to gradually share the parameters, promoting the alignments of the image and texture. Moreover, the 
adversarial masking data augmentation is extended to the 
contrastive representation learning of image and texture in a unified manner, obtaining a unified representations and further improving the \emph{ante-hoc} interpretability.} 
\label{fig:adios}
\end{figure}

Most current Medical-VQA models \citep{tmi2qcmlb,mamlffinn2017model,mevf_san_nguyen2019overcoming} mostly focus on extracting image and texture representations with dual independent encoders and grasping the interaction of vision-texture information through an extra fusion module,  which makes it challenging to learn semantic alignments. Nevertheless,  recently advanced general VQA models  \citep{li2021align,li2022blip,wang2021vlmo,chen2022utc} try to conduct a unified framework that jointly learns the cross-modal representations or constructs a unified fusion module with pre-trained features. These works motivate us to construct a unified cross-modal learning structure for Medical-VQA.
In our work, we propose the \textbf{UnICLAM} illustrated in Fig. \ref{fig:adios}, a \textbf{Un}ified and  \textbf{I}nterpretable model through the \textbf{C}ontrastive Representation \textbf{L}earning method with \textbf{A}dversarial \textbf{M}asking.
Specifically, we construct a unified dual-stream pre-training structure with a gradually soft-parameter sharing strategy for vision and texture representation alignments. 
Technically, the parameters of image and texture encoders are constrained to be close in the same space by penalizing their ${{l}_{2}}$ distances, and the higher the layer the looser the constraint.

Moreover, 
in recent work, the random mask augmentation is generally employed for obtaining superior representations in the contrastive representation learning fashion. MoCo \citep{MOCOchen2020improved}, SimCLR \citep{SIMCLRchen2020simple}, SimSiam \citep{simsiamChen_2021_CVPR} which are fundamental works in contrastive learning, employing diverse data augmentations including the random mask to generate attracted and repulsed samples, achieving remarkable performance. Some works \citep{15maskwettig2022should,feichtenhofer2022masked} also obtain promising performance with the powerful random mask exclusively.
MAE \citep{MAEhe2022masked} and  BEiT \citep{bao2021beit} achieve highly competitive results through inpainting the images occluded by random masks. 
 He \citep{Li2022ScalingLP} \emph{et al.} currently present a simple and efficient language-image pre-training method through random masks.
 However, although the random mask is widely used and effective, the representation learning under the random mask is conceptually less interpretive, as the random mask partially obscures meaningful entities
  \citep{shi2022adversarial,lisemmae}.
We hence elaborately extend the  adversarial masking augmentation strategy to the unified contrastive representation learning of vision and texture, which jointly promotes the cross-modal prediction and 
interpretation.
Concretely, while the encoder training minimizes the distance between the original and the masking samples, the masking model keeps adversarial learning to conversely maximize the distance. Through this, the adversarial masking model is optimized to generate more semantic vision and texture masks which are semantically associated for learning unified representations and improving interpretability.
Moreover, we take a further step to explore the potential \emph{ante-hoc} interpretability  (methods to incorporate the interpretability along with the model’s training process itself) of adversarial masking which is different from the \emph{post-hoc} interpretation tools  (methods to explain a previously trained model) like attention map with Grad-CAM \citep{selvaraju2017grad}.
We conduct detailed comparisons which illustrate that we have constructed an improved \emph{ante-hoc} interpretable framework with remarkable efficiency. To be specific, with the generated semantic masks of vision and text which are closely corresponding,
the unified and interpretable model boosts the precise cross-modal prediction and jointly indicates how the adversarial masking efficiently benefits the cross-modal tasks.

 For demonstrating the effectiveness of UnICLAM, we first conduct quantitative analysis on the downstream public Medical-VQA datasets including VQA-RAD \citep{vqaradlau2018dataset} and SLAKE \citep{slakeliu2021slake}. Experimental results reveal that UnICLAM achieves substantial improvements over existing $11$ state-of-the-art models. We also conduct a complicated analysis of the unified adversarial masking strategy, which further explores the potentials of \emph{ante-hoc} interpretation with considerable efficiency.
More importantly, for evaluating the performance of UnICLAM applied to real-world medical scenarios, we newly conduct a practical Medical-VQA dataset of heart failure, taking an additional discussion about the few-shot performance of the UnICLAM in diagnosing heart failure.
 The experiments reveal that our model boosts strong few-shot adaption performance by quickly adapting to practical medical applications via only limited in-context datasets.
 
Our contributions can be summarized as follows:
\begin{itemize}
\item[$\bullet$]We propose a unified and interpretable Medical-VQA structure UnICLAM, with the aid of a uniform training space where vision and texture encoders share parameters for joint interaction and a unified contrastive representation learning method with adversarial masking augmentation strategy for the semantic alignment.

\item[$\bullet$] We present an adversarial masking data augmentation strategy with unified contrastive representation learning of vision and texture that can boost the precise prediction through cross-modal alignment and further serve as \emph{ante-hoc} interpretation tools with considerable efficiency jointly.

\item[$\bullet$] Qualitative and quantitative experimental results demonstrate the outperformance of our proposed model which proves comparable prediction performance and indicative \emph{ante-hoc} interpretability of the Medical-VQA.

\item[$\bullet$] We take an additional exploration to discuss the prospect of Medical-VQA in practical heart failure diagnosing and carry out comparison experiments to verify the few-shot adaption performance of the proposed model in the newly conducted practical Medical-VQA dataset.

\end{itemize}

\section{Related Works and Preliminary}

\subsection{Medical Vision Question Answering}
Medical Vision Question Answering (Medical-VQA) is a division of the general VQA that is receiving increasing concern. Given a medical image and related clinical question, Medical-VQA model aims to generate a correct answer by analysing the vision and texture features.  Most current Medical-VQA models directly employ separate pre-trained unimodal encoders for extracting vision and texture features and adopt an extra fusion module to model the interaction between cross-modal feature, such as QC-MLB \citep{tmi2qcmlb}, MAML \citep{mamlffinn2017model}, MEVF \citep{mevf_san_nguyen2019overcoming} and MMBERT \citep{khare2021mmbert},  which suffer from the nonalignment cross-modal information.
Besides, with the independency of vision and texture encoders, other Medical-VQA works \citep{clipeslami2021does,do2021multiple,liu2021contrastive,11gong2021cross,cong2022anomaly} directly adopt the pre-trained texture encoder and only focus on vision representation learning with the image encoder  for subsequent cross-modal tasks.
Despite the above unimodal vision representation learning methods for Meical-VQA, these works indicate a tendency to learn a unified representation of vision and texture for cross-modal tasks.
 
Furthermore, current general VQA models achieve competitive results by employing a unified single-stream structure to learn aligned image and text representations, such as ALIGN \citep{alignjia2021scaling}, BLIP \citep{li2022blip}, VLMO \citep{wang2021vlmo}, and UTC \citep{chen2022utc}. 
 In addition, other works \citep{2021UNIMO,uniclyang2022unified,akbari2021vatt} concentrate on learning unified uni-modal representations through individual contrastive learning. 
 With these motivations, we propose a unified dual-stream structure with contrastive representation learning of vision and texture in a unified manner, employing a gradually soft-parameter sharing method to constrain dual heterogeneous encoders into the same close space for direct semantic interaction.

 \subsection{Contrastive Representation Learning with Random Mask Augmentation}
 Recent works have seen rapid development in contrastive representation learning by utilizing 
 diverse random augmentations to learn superior representations, such as MOCO \citep{MOCOchen2020improved}, SimCLR \citep{SIMCLRchen2020simple}, and SimSiam \citep{simsiamChen_2021_CVPR}.
 The random mask  is generally
employed as a data augmentation strategy to 
generate attracted positive samples and repulsed negative samples.
Furthermore, some works like MAE \citep{MAEhe2022masked}, SiT \citep{atito2021sit} and BEiT \citep{bao2021beit} exhibit  promising performance  with the powerful random mask strategy exclusively.
 For a further study, Chen \emph{et al.} \citep{chen2022utc} take a detailed discussion which indicates that the higher random mask rates benefit the baseline.  
 More recently, He \emph{et al.} \citep{Li2022ScalingLP} just utilize  the FLIP to improve the efficiency of both accuracy and speed by randomly masking out and removing partial image patches for training.
Actually, although random mask schemes have widespread application and are experimentally effective, the representation learning with random mask augmentation strategy still exists the deficiency of explanation, as random mask partially obscures meaningful feature entities \citep{shi2022adversarial,lisemmae}. Therefore, we extend  the adversarial masking data augmentation strategy to the unified contrastive learning of image and text that not only explicitly learns semantically aligned cross-modal representations  but also plays the role of diversified \emph{ante-hoc} 
explanation tools with efficiency. Furthermore, we take one step further to investigate the potential \emph{ante-hoc} interpretability of our model compared with the \emph{post-hoc} interpretation tools like attention map through Grad-CAM \citep{selvaraju2017grad}, which demonstrates that the unified adversarial masking module is essential in aiding splendid prediction performance and \emph{ante-hoc} interpretability simultaneously. 

 
\section{Methodology}
\label{sec:proposed method}
In this section, we propose the UnICLAM, which is a  unified and interpretable Medical-VQA model based on contrastive representation learning with adversarial masking for vision and texture. Fig. \ref{fig:UnIclam} illustrates the overview of pre-training and the fine-tune structure.

\subsection{Unified Cross-modal Pre-train Structure with Gradually Soft-parameter Sharing Strategy}
\label{sec:unified}
\begin{figure*}[htbp]
\centering
\includegraphics[width=1\linewidth]{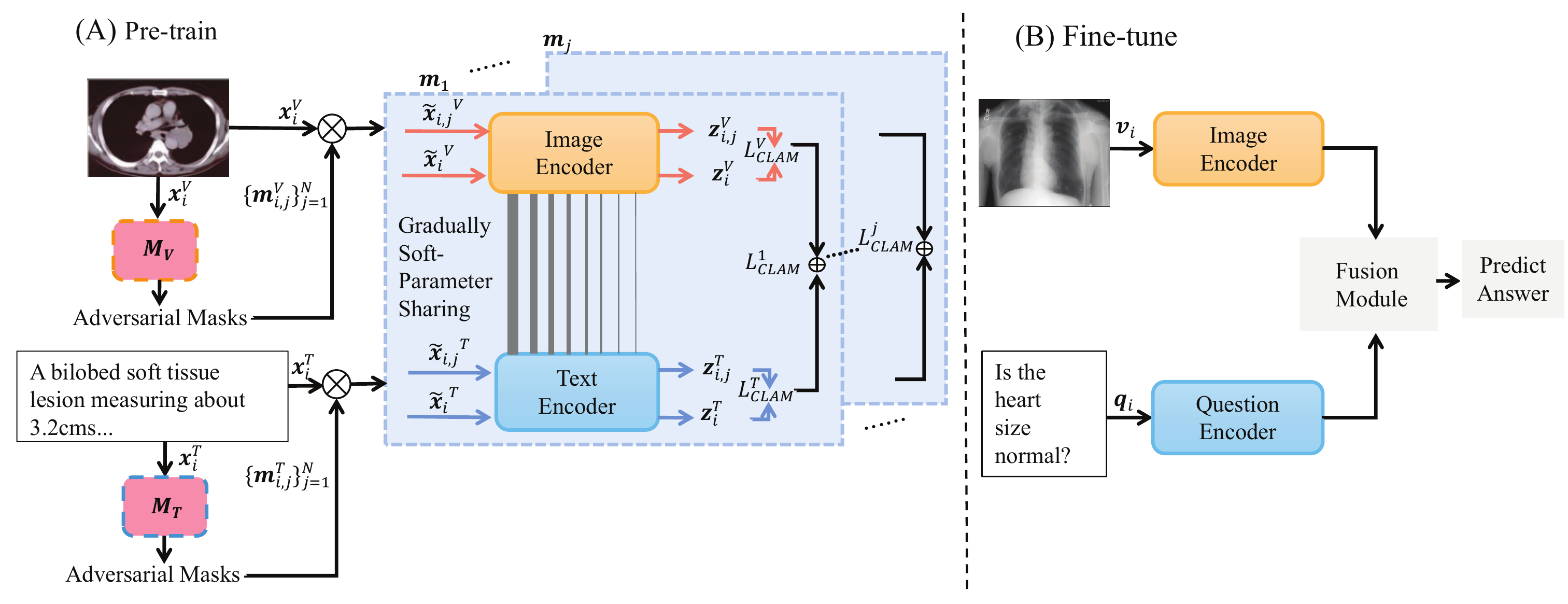}
\caption{The network structure of UnICLAM. $(A)$ Pre-train structure. The proposed UnICLAM is a dual-stream structure, which achieves cross-modal alignment by gradually soft-parameter sharing of image and texture encoders. Moreover, combined with unified contrastive representation learning,
the adversarial masking model is applied to generate senmantic masks with adversarial  learning, promoting to improve the \emph{ante-hoc} interpretablity of Medical-VQA model collaboratively.
$(B)$ Fine-tune structure with pre-trained image and question encoder. The detail introduction is represented in Section \ref{sec:proposed method}.}
\vspace{-0.3cm}
\label{fig:UnIclam}
\end{figure*}

The overall pre-training procedure is presented in Fig. \ref{fig:UnIclam} (A), we pre-train UnICLAM using a dual-stream architecture, with $Vi{{T}_{32}}$ image encoder back-end $E{_V}$ and the transformer-based text encoder $E{_T}$ initial from the BERT \citep{devlin2018bert}. For overcoming the limitation of the two resided encoders that cannot model the vital cross-modal interactions, we define a gradually soft-parameter sharing strategy to construct a closed same space where the heterogeneous gap is narrowed for vision and texture encoders. 

Given the proposed dual-stream structure, the image encoder and text encoder have same section of $K$-layer transformer units. 
The image and texture encoders are constrained into a same space by penalizing their ${{l}_{2}}$ distance. We iterate the gradually soft-parameter sharing strategy from the 1st to the ($K$-1)-th layer, promoting the cross-modal alignment  directly. 
Accordingly, the gradually soft-parameter sharing loss term ${{L}_{gs}}$  is as follows:

\begin{equation}
\label{eqn:gradua}
{{{L}_{gs}}(\partial )=\sum\limits_{k=1}^{K-1}{({{e}^{\frac{K-k}{K}}}-1)}||\partial _{k}^{E{_{V}}}-\partial _{k}^{E{_{T}}}|{{|}^{2}}}
\end{equation}
Where $\partial _{k}^{E{_{V}}}$,  $_{k}^{E{_{T}}}$ represent  the vision and texture parameter at the $k$-th transformer layers respectively. 

\subsection{Unified Image and Texture Contrastive Representation Learning}
\label{sec:cons}




First, for the contrastive representation learning, given the batch samples of $n$ input features $\left\{ {{{\boldsymbol{x}_i}}} \right\}_{i = 1}^n$, $\tilde{\boldsymbol{x}}_i$ and $\tilde{\boldsymbol{x}}_j$ 

are two diverse views which are produced by two data augmentations. The same encoder $E$ is adopted to extract the $2n$ representations from the two data augmentation views, obtaining the $\boldsymbol{z}_{i}$ and $\boldsymbol{z}_{j}$  respectively. For each augmented sample, the other augmented sample conducted from the same source defined the positive, and the remaining $(2n-2)$ samples are the negatives. The 
contrastive loss is stated as follows:

\begin{equation}
    \label{eqn:cl}
        {{L}_{CL}}(\boldsymbol{x},E)=-\log \frac{\exp (S({\boldsymbol{z}}_i,{\boldsymbol{z}}_j)/\tau )}{\mathop{\sum\nolimits_{k = 1}^{2n} {{{\rm I}_{[k \ne i]}}}{\exp (S(\boldsymbol{z}_{i},\boldsymbol{z}_{k})/\tau)} } }\
\end{equation}
Where $S$ denotes the cosine similarity, $\tau$ denotes the temperature parameter and $\rm I$ is the indicator.  

Then we extend the $N$ semantic masks 
$\left\{ {{{\boldsymbol{m}_{i,j}}}} \right\}_{j = 1}^N$ generated by the adversarial masking augmentation model $M$ to the vision and texture contrastive representation learning rather than the random masks, aiming to promote cross-modal prediction performance and improve the \emph{ante-hoc} interpretability simultaneously.
The detailed methodology of the vision adversarial $N$-masking model ${M}_{V}$ and likewise the texture ${M}_{T}$  are conducted in the next section.
Given these, the unified CLAM loss over single adversarial mask  ${\boldsymbol{m}_{i,j}}$ is proposed:

\begin{equation}
    \label{eqn:clam}
    \begin{split}
      \resizebox{0.9\hsize}{!}{${{L}_{CLAM}}(\boldsymbol{x},E,M)=
    -\log \frac{\exp (S(E(\boldsymbol{x}_{i}\odot M(\boldsymbol{x}_{i})),E(\boldsymbol{x}_{i}))/\tau)}{\sum\limits_{k\ne i}{\exp (S(E(\boldsymbol{x}_{i}\odot M(\boldsymbol{x}_{i})),E(\boldsymbol{x}_{k}))/\tau)}} $}
    \end{split}
\end{equation}

\begin{equation}
\begin{aligned}
          & L_{CLAM}^{V}={{L}_{CLAM}}(\boldsymbol{x}^V,E{_{V}},{{M}_{V}}) \\ 
 & L_{CLAM}^{T}={{L}_{CLAM}}(\boldsymbol{x}^T,E{_{T}},{{M}_{T}}) \\ 
\end{aligned}
\end{equation}
Where $M(\boldsymbol{x}_{i})$  is the semantic mask ${\boldsymbol{m}_{i,j}}$ generated by adversarial masking model $M$, and $\boldsymbol{x}_{i}$ denotes the original input sample. 
$L_{CLAM}^{V}$ and $L_{CLAM}^{T}$ stands for the CLAM loss of the image features $\left\{ {\boldsymbol{x}_i^V} \right\}_{i = 1}^n$ and the texture features 
 $\left\{ {\boldsymbol{x}_i^T} \right\}_{i = 1}^n$ respectively.


\subsection{Adversarial Masking Augmentation Model for Image and Texture}
\label{adver}
In general, since each input medical image is composed of several visual entities and the related question contains different word tokens, we generate $N$ masks of each input vision features  $\boldsymbol{x}_i^V$ and texture features  $\boldsymbol{x}_i^T$ through the same adversarial masking data augmentation method with the masking model ${M}_{V}$ and ${M}_{T}$. 
Notably, the ${M}_{V}$ and ${M}_{T}$ all consist of a learnable neural network and the softmax layer respectively. 
The generated image and texture semantic masks are closely semantically associated to learn aligned cross-modal representations and improve the \emph{ante-hoc} interpretability.

For a continuous image feature  $\boldsymbol{x}_i^V$, we directly embed it into the vision adversarial masking model ${M}_V$. 
Since the question feature $\boldsymbol{x}_i^T$ are discrete, we first encode the  $q$ question tokens into an embedding matrix through a sentence embedding layer ${{f}_{emb}}$ for adversarial learning, where 
${\boldsymbol{x}_i^T}:{{\mathbb{R}}^{q}}\mapsto {{\mathbb{R}}^{q\times h}}$
, $h$ is the embedding dimension.  We can write the generation of $N$ image masks $\left\{ {{{\bm{m}_{i,j}^{V}}}} \right\}_{j = 1}^N$ and texture masks $ \left\{ {{{\boldsymbol{m}_{i,j}^{T}}}} \right\}_{j = 1}^N$ as:
\begin{equation}
\left\{ {{{\boldsymbol{m}_{i,j}^{V}}}} \right\}_{j = 1}^N={{f}_{head}}({{M}_{V}}({\boldsymbol{x}_i^V}))   
\end{equation}
\begin{equation}
   \left\{ {{{\boldsymbol{m}_{i,j}^{T}}}} \right\}_{j = 1}^N={{f}_{head}({M}_{T}}({{f}_{emb}}({\boldsymbol{x}_i^T}))) 
\end{equation}
Where the ${{f}_{head}}$ represents a convolution layer 
for mapping the output of the masking model to $N$ masks. The ${\boldsymbol{x}_i^V}:{{\mathbb{R}}^{w\times h}}$, $w$ represent for the width of the image and $h$ is the height.

For the $N$ generated semantic masks $\left\{ {{{\boldsymbol{m}_{i,j}^{V}}}} \right\}_{j = 1}^N$ and $\left\{ {{{\boldsymbol{m}_{i,j}^{T}}}} \right\}_{j = 1}^N$, we employ them into $N$ forward pass of the image and texture contrastive representation learning. 
While the image encoder $E{_V}$ and texture encoders $E{_T}$ keep training with the unified contrastive learning to minimize the distance between the original feature and the masking feature, the masking model $M$ keeps adversarial learning to conversely maximize the distance. Generally, the adversarial masking model $M$   across $N$ masks is optimized as follows:

\begin{equation}
\label{loss_all}
{{E}^{'}},{{M}^{'}}=\arg \underset{E}{\mathop{\min }}\,\underset{M}{\mathop{\max }}\,\frac{1}{N}\sum\limits_{j=1}^{N}{L_{CLAM}^{j}(\boldsymbol{x},E,M)}
\end{equation}
\begin{equation}
\label{eqn:loss_clam}
{{L}_{CLAM}}(\boldsymbol{x},E,M)=L_{CLAM}^{V}+L_{CLAM}^{T}
\end{equation}

After the adversarial learning of masking model $M$, the complete pre-training objective of UnICLAM is optimized to minimize the $ {L}_{UnICLAM}$  which can be denoted by:
\begin{equation}
\label{lossuniclam}
    \resizebox{0.9\hsize}{!}{${{{L}_{UnICLAM}}=\beta*L_{CLAM}^{V}+(1-\beta  )*L_{CLAM}^{T}+{\lambda  * {L}_{gs}}(\partial )}$}
\end{equation}
Where $\beta$,$\lambda$ are the hyperparameter for balancing  loss terms.

\subsection{Fine-tune in Medical-VQA}

The Medical-VQA is a multi-class classification task and we implement our pre-trained image encoder $E{{n}_{V}}$ and texture encoders $E{{n}_{T}}$ upon it as shown in Fig. \ref{fig:UnIclam} (B). 
Given a fine-tune dataset $D$ which contains medical image set ${\boldsymbol{v}_{i}}\in V$, the related clinical question ${\boldsymbol{q}_{i}}\in Q$ and the answer ${{a}_{i}}\in A$.  
We aim to predict the answer by optimizing the predicting formulation $f$, which is formulated as:

\begin{equation}
    {{f}_{\mu }}={{C}_{\mu }}({{M}_{\mu }}(E{_{V}}_{\mu }({\boldsymbol{v}_{i}}),E{_{T}}_{\mu }({\boldsymbol{q}_{i}})))
\end{equation}
Where ${{M}_{\mu }}$ denotes the fusion network,
${{C}_{\mu }}$ is the classification. $\mu $ is a learnable parameter that is optimized as follows:
\begin{equation}
\mu =\underset{\mu }{\mathop{\arg \max }}\,\sum\limits_{i=1}^{D}{\log p(}{{a}_{i}};{{f}_{\mu }}({\boldsymbol{v}_{i}},{\boldsymbol{q}_{i}}))
\end{equation}

\section{Datasets and Implementation details}
\subsection{Datasets for Medical-VQA}
\subsubsection{Pre-train dataset}
Radiology Objects in COntext (ROCO) datasets \citep{ROCOpelka2018radiology} which is the only large public medical cross-modal dataset that provides over $81,000$ radiology images with diverse imaging modalities. The various image-text pairs are captured from \emph{PubMed} articles, the text is chosen from the relative captions, providing rich descriptive information about the medical image.
\subsubsection{Fine-tune dataset}
VQA-RAD \citep{vqaradlau2018dataset} is a radiology-specific dataset most commonly used in Medical-VQA, which includes $315$ radiological images and $3,515$ question-and-answer samples uniformly distributed over the head, chest, and abdomen. Medical images are selected from a balanced sample from \emph{MedPix®}, and the questions are posed by clinicians and annotated with answers. There are 11 categories of clinical problems, including anomalies, attributes, counting \emph{etc}. Depending on the answer type, questions are also divided into closed-ended questions with a "yes/no" answer and open-ended questions with no fixed form answer. The closed-ended and open-ended questions are all manually labeled with an right answer. The VQA-RAD contains $458$ kinds of answers in total.
In our model, we used the same data partitioning method as in [30], with 3064 question-and-answer training samples and 451 testing samples.

SLAKE \citep{slakeliu2021slake} is a abundant Medical-VQA dataset with semantic labels and a medical knowledge repository. It includes $642$ radiological images and $7,033$ question-and-answer samples distributed over $11$ diseases and $39$ organs, including 10 clinical problems such as modality, position, etc. Medical images were selected from three open-source datasets \citep{75simpson2019large,88fukui2016multimodal,40kavur2021chaos} and labeled by medical experts. 
In our experiments, following the original way of data splitting, 70$\%$ of question-and-answer pairs are  train-set and 30$\%$ are test-set.

\begin{table*}
	\centering
	\caption{The comparison of experimental results between the UnICLAM and the state-of-the-art method on VQA-RAD and SLAKE public benchmarks respectively. $*$ indicates our re-implemented result, including the mean accuracy and standard deviation by 10 runs under 10 different seeds.} 
	\label{VQAresult}
	\scalebox{1.}{
	\setlength{\tabcolsep}{6pt}
	 \renewcommand{\arraystretch}{1.2}
\begin{tabular}{l|cccccccc}
\cline{1-7}
\multirow{2}{*}{\textbf{Methods}} & \multicolumn{3}{c}{\textbf{VQA-RAD }}                                            & \multicolumn{3}{c}{\textbf{SLAKE }}                                              & \textbf{} & \textbf{} \\ \cline{2-7}
                                  & \multicolumn{1}{c}{\textbf{Open}} & \multicolumn{1}{c}{\textbf{Closed}} & \multicolumn{1}{c}{\textbf{Overall}} & \multicolumn{1}{c}{\textbf{Open}} & \multicolumn{1}{c}{\textbf{Closed}} & \multicolumn{1}{c}{\textbf{Overall}} &           &           \\ \cline{1-7}
MFB \citep{mfbyu2017multi}                               & 14.5                     & 74.3                       & 50.6                    & 72.2                     & 75.0                       & 73.3                    &           &           \\
SAN \citep{sanyang2016stacked}                               & 31.3                     & 69.5                       & 54.3                    & 74.0                     & 79.1                       & 76.0                    &           &           \\
BAN \citep{kim2018bilinear}                               & 37.4                     & 72.1                       & 58.3                    & 74.6                     & 79.1                       & 76.3                    &           &           \\
MAML+SAN \citep{finn2017model}                          & 38.2                    & 69.7                       & 57.1                    & /                     & /                       & /                   &           &           \\
MAML+BAN \citep{finn2017model}                          & 40.1                    & 72.4                       & 60.7                    & /                     & /                       & /                   &           &           \\
MEVF+SAN(*) \citep{mevf_san_nguyen2019overcoming}                          & 49.2$_{\pm 1.8}\%$                     & 73.9$_{\pm 1.7}\%$                       & 64.1$_{\pm 1.8}\%$                    & 75.3$_{\pm 1.9}\%$                     & 78.4$_{\pm 1.4}\%$                       & 76.5$_{\pm 1.6}\%$                    &           &           \\
MEVF+BAN(*) \citep{mevf_san_nguyen2019overcoming}                          & 49.2$_{\pm 1.9}\%$                     & 77.2$_{\pm 1.6}\%$                      & 66.1$_{\pm 1.7}\%$                    & 77.8$_{\pm 1.6}\%$                     & 79.8$_{\pm 1.3}\%$                      & 78.6$_{\pm 1.5}\%$                    &           &           \\
VQAMix \citep{gong2022vqamix}                            & 56.6                     & 79.6                       & 70.4                    & /                        & /                          & /                       &           &                     \\ \cline{1-7}
MMBERT(*) \citep{khare2021mmbert}                         & 58.3$_{\pm 1.6}\%$                     & 76.9$_{\pm 1.4}\%$                       & 66.9$_{\pm 1.4}\%$                    & /                        & /                          & /                       &           &           \\
CRPD+BAN \citep{liu2021contrastive}                          & 52.5                   & 77.9                       & 67.8                    & 79.5                     & 83.4                       & 81.1                   &           &\\ 
PubMedCLIP-ViT-B+QCR(*) \citep{clipeslami2021does}           & {58.0}$_{\pm 1.7}\%$                      & {79.6$_{\pm 1.4}\%$}                           & {71.1}$_{\pm 1.1}\%$                    & {78.2$_{\pm 1.8}\%$}                     & {82.6$_{\pm 1.2}\%$}                       & {80.1$_{\pm 1.5}\%$}                    &           &           \\ \cline{1-7}
\textbf{Ours}                              & \textbf{59.8}$_{\pm 1.4}\%$                 & \textbf{82.6$_{\pm 1.0}\%$}                       & \textbf{73.2}$_{\pm 0.9}\%$                    & \textbf{81.1}$_{\pm 1.5}\%$                      & \textbf{85.7}$_{\pm 0.8}\%$                        & \textbf{83.1}$_{\pm 1.1}\%$                     &           &           \\ \cline{1-7}
\end{tabular}}
\end{table*}

\subsection{Implementation Details}
The UnICLAM is applied with Pytorch library and trained with  6 NVIDIA TITAN 24 GB Xp GPUs.
\subsubsection{Pre-taining}
We employ the SimCLR \citep{SIMCLRchen2020simple} as the contrastive representation learning baseline and adopt the 
U-Net and $3$-layer shallow transformer as the adversarial masking model  ${{M}_{T}}$ and ${{M}_{V}}$. The number of image and text adversarial masks is set to  $4$ and $2$ respectively. 
For the implementation, We take the image crops to $224\times 224$ as input. Each text is trimmed to 32-words, and the dimension of the sentence embedding layer ${{f}_{head}}$ is $512$.
The number $n$ of transformer layers in vision and texture encoder is set to $12$ with the hidden layer set to 1024 dimensions. The convolution layer ${{f}_{head}}$ with   $1\times 1$ kernel is utilized to produce adversarial masks. 
The hyperparameter $\lambda$ and $\beta$ of loss terms in Eqn. \ref{lossuniclam} are set to $1{{e}^{-3}}$ and $0.3$ respectively.
The UnICLAM is trained for $40$ epochs using $64$ batch size. The encoders of the vision and the texture are randomly initialized and trained with Adam \citep{2014Adam} optimizer whose initial learning rate and weight decay  set to $1{{e}^{-5}}$ and  $5{{e}^{-4}}$ respectively.  

\subsubsection{Fine-tune}
For fine-tuning, the dual-encoders are initialized with the pre-trained encoders $E{_V}$ and $E{_T}$. We adopt the bilinear attention-based network (BAN \citep{kim2018bilinear}) as the fusion module  ${{M}_{\mu }}$. We crop the images to $224\times 224$, and the questions are trimmed to 12 tokens and then embedded into 300-dimensions through GloVe \citep{pennington2014glove} for both VQA-RAD and SLAKE. 
The experiment results of mean accuracy and standard deviations are computed by 10 different seeds \emph{e.t.} $100-1000$.

\subsection{Comparison with State-of-the-art Methods}
The UnICLAM is compared with the $11$ current state-of-the-art models \citep{mfbyu2017multi,mevf_san_nguyen2019overcoming,sanyang2016stacked,kim2018bilinear,crzhan2020medical,khare2021mmbert,gong2022vqamix,clipeslami2021does}. We also re-implement four baselines with same employment of datasets. The brief introduction of above state-of-the-art models is shown below.
\begin{itemize}
\item[$\bullet$] MFB \citep{mfbyu2017multi} SAN \citep{sanyang2016stacked} 
BAN \citep{kim2018bilinear} models are general VQA frameworks employed in medical domain, residing image-text encoders in separate spaces and then modeling the cross-modal features with a simple fusion model.

\item[$\bullet$] MAML \citep{mevf_san_nguyen2019overcoming} utilizes the meta-learning methods to pre-train the vision encoder for a quick adaption to new cross-modal tasks. MAML+SAN and MAML+BAN adopt the Stacked Attention Networks (SAN) and Bilinear Attention Network (BAN) as the fusion module.

\item[$\bullet$] MEVF \citep{mevf_san_nguyen2019overcoming} is the baseline that pre-trains the vision encoder with the combination of convolutional denoising auto-encoder and meta-learning.
MEVF+SAN, MEVF+BAN adopt the SAN and BAN to fuse pre-trained vision representations and texural features. 

\item[$\bullet$] 
CRPD \citep{liu2021contrastive} leverages medical images to train and distill the  general image encoder for Medical-VQA, employing the contrastive representation learning with 
random mask augmentation strategy. 
CRPD+BAN combines with the BAN attention mechanism for cross-modal fusion.

\item[$\bullet$] MMBERT \citep{khare2021mmbert} adopts the BERT \citep{devlin2018bert} to grasp the interaction of the images and texture through the Masked Language Modeling tasks with random mask strategy.

\item[$\bullet$] VQAMix \citep{gong2022vqamix} claims to mitigate data limitation in medical VQA through the conditional-mixed labels data augmentation strategy with great interpretability. 

\item[$\bullet$] PubMedCLIP-ViT-B+QCR \citep{clipeslami2021does} applies the Contrastive Language-Image Pre-training(CLIP) with random augmentation strategy to better extract vision representations for Medical-VQA, whereas the question conditional reasoning module is used to improve reasoning ability.
\end{itemize}

\section{Experiments and results}
\label{sec:guidelines}

\subsection{Experimental Results}
\subsubsection{Comparison with State-of-the-art.}
The comparison in Table \ref{VQAresult} illustrates that our proposed UnICLAM is superior to the state-of-the-art methods, in particular, the UnICLAM yields 73.2$\%$  mean accuracy in VQA-RAD and 83.1$\%$ in SLAKE respectively.  
Besides, as the left part of Table \ref{VQAresult} shows, our proposed UnICLAM obviously outperforms the baseline models MFB \citep{mfbyu2017multi}, SAN \citep{sanyang2016stacked} and BAN \citep{kim2018bilinear}. The improvements over MAML+BAN \citep{mamlffinn2017model} and MEVF+BAN \citep{mevf_san_nguyen2019overcoming} which all adopt the same fusion module BAN \citep{kim2018bilinear} are $12.5\%$ and $7.1\%$ overall accuracy in VQA-RAD, respectively.
Explicitly, the UnICLAM considerably surpasses the VQAMix \citep{gong2022vqamix} which proposes a simple yet effective data augmentation method by $2.8\%$ overall performance gain, indicating that the proposed model may has superiority in learning unified vision-texture representations for Medical-VQA.
Moreover, our model outperforms the advanced approaches MMBERT \citep{khare2021mmbert}, CPRD+BAN \citep{liu2021contrastive}, and PubMedCLIP \citep{clipeslami2021does} which all employ random augmentation strategy by up to $6.3\%$, $5.4\%$, and $2.1\%$ overall accuracy.
Also, we observe that our proposed UnICLAM can considerably outperform compared with the  \citep{khare2021mmbert,clipeslami2021does} which pre-train with image-caption pairs. A possible explanation could be that, as expected, the proposed model within a unified and interpretable cross-modal structure acts a significant role in the learning of vision and texture representations for cross-modal alignment.

Besides, the experiments results in the right section of Table \ref{VQAresult} illustrate that UnICLAM still outperforms the baseline models \citep{mfbyu2017multi,sanyang2016stacked,kim2018bilinear,crzhan2020medical} in SLAKE dataset. 
It also can be witnessed that although the strong baseline MEVF+BAN  \citep{mevf_san_nguyen2019overcoming} brings $2.3\%$ overall performance gain on SLAKE, which also pre-trains vision encoder on medical images and employs the same fusion structure BAN \citep{kim2018bilinear} as ours, it is far from enough compared to the $6.8\%$ overall performance gain by ours. In particular, although  PubMedCLIP-ViT-B+QCR \citep{clipeslami2021does} enhances the vision representations by pre-training vision encoder with image-caption pairs and utilizing the question conditional reasoning module, the proposed UnICLAM dramatically improves the overall accuracy by $3\%$ (closed-ended improved  $3.1\%$ and open-ended improved $2.9\%$). This demonstrates that our model has the superiority to achieve precise prediction with unified representations for Medical-VQA.

\begin{table}

	\centering
	\caption{Ablation study on VQA-RAD with the respect to the experiments results. $"Aug"$ means the data augmentation strategy in contrastive representation leaning. $"RM"$:random mask,  $"AM"$:Adversarial Masking.   $"GSPS"$: Gradually Soft-Parameter Sharing strategy. $"UnI":$ Unified vision and texture contrastive representation learning.}  
	\label{ablation}
      \setlength{\tabcolsep}{4pt}
	 \renewcommand{\arraystretch}{1.5}
	\scalebox{0.9}{

\begin{tabular}{cccccccccc}
\hline
\multirow{2}{*}{\textbf{Index}} & \multicolumn{2}{c}{\textbf{Aug}} & \multirow{2}{*}{\textbf{GSPS}} & \multirow{2}{*}{\textbf{UnI}} & \multicolumn{3}{c}{\textbf{VQA-RAD}}                \\ \cline{2-3} \cline{6-8} 
                                & \textbf{RM}     & \textbf{AM}    &                                &                               & \textbf{Open} & \textbf{Closed} & \textbf{Overall} \\ \hline
1                               &  {\checkmark  }                & ×              & ×                              & {\checkmark  }                              & 57.6$_{\pm 1.6}\%$          & 77.9$_{\pm 1.1}\%$            & 70.1$_{\pm 1.2}\%$             \\
2                               &  {\checkmark  }               & ×              &  {\checkmark  }                               & {\checkmark  }                              & 58.6$_{\pm 1.2}\%$          & 80.1$_{\pm 1.4}\%$            & 71.4$_{\pm 1.3}\%$             \\
3                               & ×               &  {\checkmark  }               &  {\checkmark  }                               & ×                             & 59.5$_{\pm 1.1}\%$          & 81.8$_{\pm 1.2}\%$            & 72.7$_{\pm 1.0}\%$             \\
4                               & ×               &  {\checkmark  }               & {\checkmark  }                               &{\checkmark  }                              & 59.8$_{\pm 1.4}\%$        & 82.6$_{\pm 1.0}\%$            & 73.2$_{\pm 0.9}\%$            \\ 
\hline

\end{tabular}}
\end{table}

\subsubsection{Ablation study}
Table \ref{ablation} demonstrates the ablation study which verifies the effectivity of each devised method, including the \textbf{G}radually \textbf{S}oft-\textbf{P}arameter \textbf{S}haring strategy(\textbf{GSPS}) and the \textbf{UnI}fied contrastive representation learning method (\textbf{UnI}) with \textbf{A}dversarial \textbf{M}asking (\textbf{AM}).

For index $1$, we take a  dual-stream model which equipped with hard-parameter sharing strategy and pre-trained through the unified vision and texture contrastive representation learning with random mask data augmentation as the baseline. As shown in index $2$, the baseline model equipped with a gradually soft-parameter sharing strategy obverses 1.3$\%$ overall accuracy increases on VQA-RAD (the open-ended and closed-ended achieve an absolute accuracy gain of 1$\%$ and 2.2$\%$ respectively). This result demonstrates that the gradually soft-parameter sharing strategy can bring performance gains through aligning unified representations of the vision and texture directly in the same close space.

\begin{table}
	\caption{Mean Overall accuracy(and standard deviation) of the number of vision ${N_v}$ and texture adversarial masks ${N_t}$.}  
	\label{22}
 	\setlength{\tabcolsep}{2pt}
	 \renewcommand{\arraystretch}{1.5}
	\scalebox{0.85}{
\begin{tabular}{cllllll}
\hline
\multirow{3}{*}{\textbf{Overall}}    & \multicolumn{6}{c}{\textbf{Number of Adversarial Masks}}     \\ \cline{2-7} 
                            & ${N_v}$=3  & ${N_v}$=4  & ${N_v}$=5  & ${N_v}$=3  & ${N_v}$=4  & ${N_v}$=5  \\
                            & ${N_t}$=2  & ${N_t}$=2  & ${N_t}$=2  & ${N_t}$=3  & ${N_t}$=3  & ${N_t}$=3  \\ \hline
\multicolumn{1}{c}{\textbf{VQA-RAD}} & 72.6$_{\pm 1.0}\%$ & \textbf{73.2}$_{\pm 0.9}{\%}$ & 72.8$_{\pm 1.1}\%$ & 72.3$_{\pm 1.2}\%$ & 72.6$_{\pm 0.8}\%$ & 72.4$_{\pm 1.3}\%$ \\
\multicolumn{1}{c}{\textbf{SLAKE}}   & 82.3$_{\pm 1.3}\%$ & \textbf{83.1}$_{\pm 1.1}{\%}$ & 82.5$_{\pm 1.0}\%$ & 82.1$_{\pm 1.4}\%$ & 82.7$_{\pm 0.9}\%$ & 82.3$_{\pm 1.4}\%$ \\ \hline
\end{tabular}}
\end{table}

\begin{figure}[htbp]
\centering
\includegraphics[width=1\linewidth]{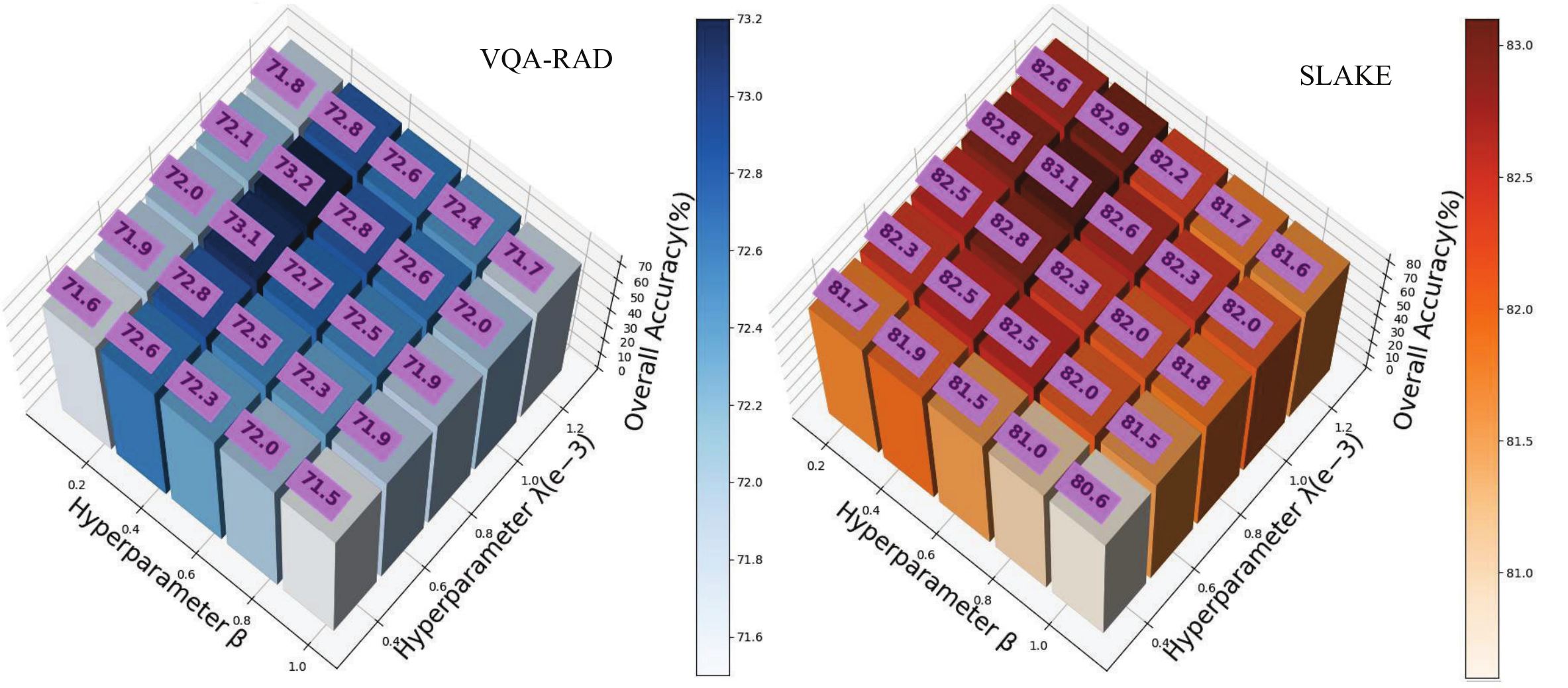}
\caption{Evaluation results of hyperparameters $\lambda$ and $\beta$ for UnICLAM loss  terms on VQA-RAD and SLAKE datasets.}
\vspace{-0.3cm}
\label{fig:hyp}
\end{figure}

In addition, within the same model architecture, we swap out the unified contrast learning method with the sequential contrastive representation learning method. The vision and question encoders only acquire their individual unimodal representations as a result of this independent training.
the comparison between the $3$rd and $4$th index shows that the unified vision and texture contrastive representation learning method has superiority in obtaining unified cross-modal representations for Medical-VQA, bringing $0.5$$\%$ accuracy improvements over the individual contrastive learning method.
Furthermore, in index $4$, replacing the random mask augmentation with the adversarial masking data augmentation strategy, it can be seen that the adversarial masking data augmentation strategy improves the mean overall accuracy from $71.4$$\%$ to $73.2$$\%$ (the closed-ended accuracy significantly improved 2.5$\%$ and the open-ended accuracy improved 1.2$\%$),  which demonstrates that the adversarial learning of masks benefits unified semantic representations learning of vision and texture and boosts the precise prediction performance. 

Accordingly, the baseline equipped with a gradually soft-parameter sharing method and the unified  contrastive representation learning strategy with adversarial masking brings obvious improvement by up to $3.1\%$ on overall accuracy of VQA-RAD dataset, suggesting that our method have grasped unified semantic representation of vision and texture as well as the alignment of cross-modal features for Medical-VQA.

\subsubsection{Influence of Adversarial Masks in Different Numbers}
For investigating the number of adversarial masks impact on the performance, we evaluate the UnICLAM with different numbers of vision and texture adversarial masks. It is obviously observed from Table \ref{22} that using $4$ image adversarial masks and $2$ texture adversarial masks can achieve the best results. Crucially, since medical images contain multiple vision entities, the $3$ generated vision masks may ignore different levels of semantic details (e.g. different objects and backgrounds), and likewise $5$ vision masks may result in an interference with redundancy masking. Meanwhile, for questions that relatively focus on the aim of the main object, the $2$ generated texture masks can effectively cover enough semantic keywords which work in concert with vision mask entities. 

\begin{figure*}
\centering
\centerline{\includegraphics[width=0.99\linewidth]{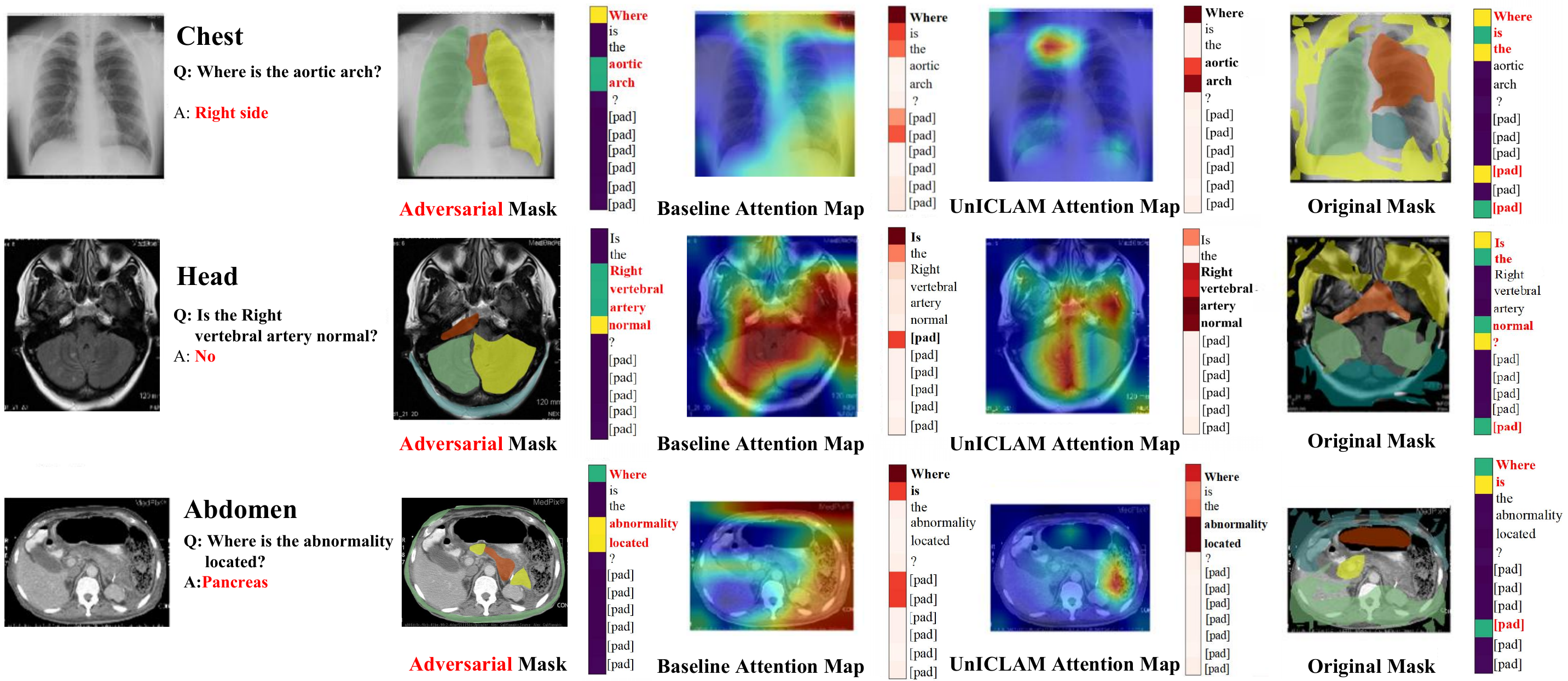}}
\caption{The interpretability comparison. These examples are selected from the VQA-RAD dataset, including an X-ray of chest, an MRI of head, and a CT of abdomen organs. First column: input image and question-answer pairs; Second column: the adversarial masking employed for occlusion of the input vision and text features. Third column: the attention of baseline model. Fourth column: the attention map of the proposed UnICLAM. The last column: the original mask employed for random occlusion of the image and question features without adversarial learning. }
\label{fig:Adversarial Masking}
\end{figure*}

\subsubsection{Influence of Hyperparameters in Loss Terms}

We investigate the effect of two hyperparameters $\lambda$ and $\beta$ in Eqn. \ref{lossuniclam} on VQA-RAD and SLAKE datasets, the experiments are summarized in Fig. \ref{fig:hyp}.
We assign the hyperparameters $\beta$ from $0.1$ to $0.9$ with step $0.2$  
and $\lambda$ from $0.4{e^{-3}}$ to $1.2{e^{ - 3}}$ with step $0.2{e^{-3}}$.
The proposed model achieves the best result with $\beta=0.3$ and $\lambda={e^{-3}}$ respectively. The mean accuracy presented by the two bar charts is insensitive to the hyperparameters $\lambda$ and $\beta$ setting, indicating the robustness of the proposed strategies with stable results.

\subsection{Analysis on Ante-hoc 
 Interpretability}
\label{qua}
As shown in Fig. \ref{fig:Adversarial Masking}, we conduct a complicated analysis to illustrate the \emph{ante-hoc} interpretability of UnICLAM and take a further step to figure out how the adversarial masking helps in answering medical images-questions in detail. 
The comparison imitates the process of predictions in four different ways, including the adversarial mask,  the baseline attention map, the UnICLAM attention map, and the original mask without adversarial learning, applying to vision and question features respectively.
 Notably, the attention map represents the model concerned with each section in an image and each word in the question, the darker color indicates that more attention is paid. The section of different colors represents different semantic masks generated by the adversarial masking module for occluding input images and questions. 

\begin{table}
	\centering
	\caption{The comparison of the explanation time per instance between the attention maps and the proposed adversarial masking.}  
	\label{time}
  	\setlength{\tabcolsep}{8pt}
	 \renewcommand{\arraystretch}{1.2}
	\scalebox{1}{
\begin{tabular}{cclcl}
\hline 
	\setlength{\tabcolsep}{6pt}
	 \renewcommand{\arraystretch}{1.3}
\multirow{2}{*}{\textbf{Methods}} & 
\multicolumn{4}{c}{\textbf{Explanation Time Per Instance}}                 \\ \cline{2-5} 
                         & \multicolumn{2}{c}{\textbf{Vision}} & \multicolumn{2}{c}{\textbf{Question}} \\ \hline
Attention Map            & \multicolumn{2}{c}{0.08$_{\pm 0.05}$s}     & \multicolumn{2}{c}{0.05$_{\pm 0.03}$s}         \\
Adversarial Mask                & \multicolumn{2}{c}{0.02$_{\pm 0.01}$s}     & \multicolumn{2}{c}{0.01$_{\pm 0.01}$s}         \\ \hline   
\end{tabular}}
\end{table}

More specifically, we select three examples from each organ included in the VQA-RAD dataset for reliability. For the X-ray of the chest organ (example $1$), 
the adversarial masking precisely focuses the keywords which are \emph{"Where, aortic arch"} to directly indicate the aim and the key subject rather than the meaningless words the baseline attention map focuses on. Meanwhile, the adversarial vision mask correctly locates semantic regions of the aortic arch as the orange mask shows while the attention map of the baseline model only highlights the coarse whole trachea as well as much irrelevant black background. 

\begin{figure*}[htbp]
\centering
\includegraphics[width=0.99\linewidth]{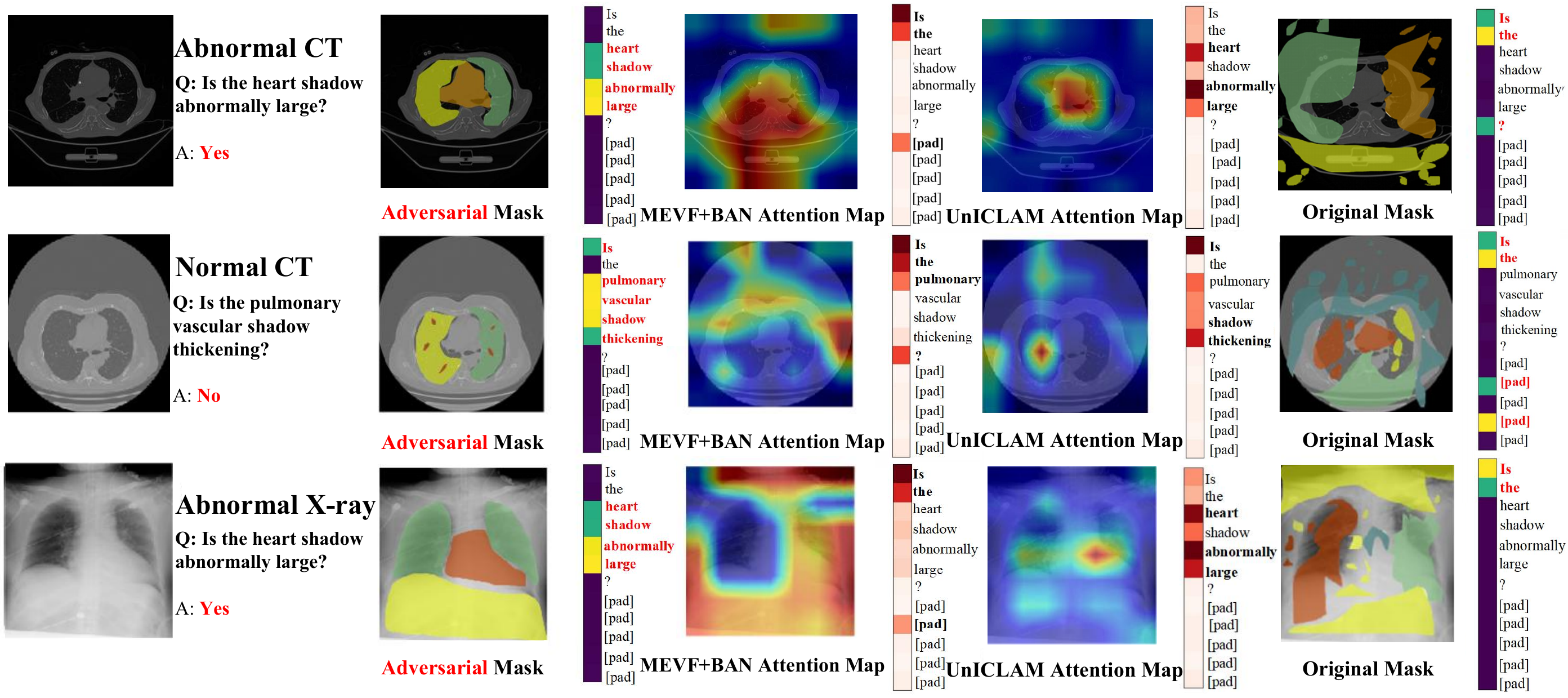}
\caption{The interpretability comparison of practical datasets. First column: practical images with question-answer pairs; Second column: adversarial mask. Third column: the attention map of MEVF+BAN model. Fourth column: the attention map of UnICLAM. The last column: original mask. }
\label{fig:practical mask}
\end{figure*}
Besides, the comparison between the adversarial mask and the attention map of  UnICLAM indicates that the generated adversarial mask has a superior indicative ability to locate precise  features and further distinctly mask out 
inhomogeneous features with the help of different colors. To be specific, 
while the attention map of UnICLAM only concentrates on the 
most critical single area like the upper part of the right thoracic cavity where is the approximate location of aortic arch, the adversarial masks not only accurately point out the outline of aortic arch (orange), but also locate other semantic feature types like bilateral lungs (green and yellow). Moreover, the original vision mask without adversarial learning in the right column is cluttered. As the yellow, green, and orange sections show, there are no particular semantic masks of visual entities.
Meanwhile, the original question mask more focuses on the irrelevant \emph{"Where, is, the"} words. On the contrary, we can witness prominent gains on the adversarial mask from the left column, after the adversarial learning, the learned question mask pays more attention to semantic words and vision regions.
More importantly, it is a breathtaking exploration that the learned vision mask (orange) and the learned question mask (\emph{aortic arch}) are closely semantically corresponding and correctly indicate the right answer in a unified manner. 




For the MRI of the head image (example $2$), compared with the whole rough brain area and weak semantics words the attention map of baseline concentrates, the adversarial masking accurately indicates the keywords 
and the relevant semantic regions around the vertebral artery as the green, yellow and orange masks shown. Furthermore, the adversarial masks trace out kinds of types including vertebral artery and bilateral cerebellum for concrete interpretation while the attention map of UnICLAM only concentrates on the rough regions around the vertebra. In addition, as the last column shows,
The learned Adversarial masks include the significant vision entity: \emph{"vertebral artery"} around the \emph{"cerebellum"} and the learned semantic words which are well indicative of each other while the original masks ignore.

We also present the effectiveness of adversarial masking strategy on abdomen (example $3$). 
It can be seen from the bottom row that adversarial masks can concentrate on more semantic words like \emph{"Where, abnormality, located"} and correctly locate the relevant image region of pancreas. 
As well, the adversarial masks precisely point out kinds of semantic features in different color like the orange and yellow area of pancreas while the attention map locates singular regions around the approximate area of pancreas.
At the same time, the learned adversarial masks of vision and question are in a joint interaction for guiding the correct answer.

\begin{table}
	\centering
	\caption{The mean accuracy ( standard deviation)  results on the practical Medical-VQA datasets. \emph{"Q1"} :\emph{"Is the heart shadow abnormally large?"}, \emph{"Q2"} stands for \emph{"Is the pulmonary vascular shadow thickening?"}. $"*"$ indicates our re-trained results.}  
	\label{pratical}
	\scalebox{0.9}{

\begin{tabular}{cccll}
\hline
\multirow{2}{*}{\textbf{Models}}      & \multirow{2}{*}{\textbf{Modality}} & \multicolumn{3}{c}{\textbf{Accuracy}}                                            \\ \cline{3-5} 
                                      &                                    & Q1                   & \multicolumn{1}{c}{Q2} & \multicolumn{1}{c}{Overall} \\ \hline
\multirow{2}{*}{MEVF+BAN(*)}             & X-rays                             & 86.9$_{\pm 0.7}\%$  & 87.8$_{\pm 0.6}\%$
&85.2$_{\pm 0.7}\%$
                            \\
                                      & CTs                                 &82.5$_{\pm 1.4}\%$
                                      &83.6$_{\pm 1.1}\%$
                                      &81.2$_{\pm 1.3}\%$                             \\
                         \multirow{2}{*}{MMBERT(*)}               & X-rays                             &87.9$_{\pm 0.9}\%$                           & 88.7$_{\pm 0.5}\%$                         & 86.8$_{\pm 0.8}\%$                                 \\
                                      & CTs                                &84.2$_{\pm 1.5}\%$                        &84.9$_{\pm 1.3}\%$                          & 82.9$_{\pm 1.4}\%$    \\
\multirow{2}{*}{PubMedCLIP(*)} & X-rays                             & \multicolumn{1}{l}{91.0$_{\pm 0.9}\%$ } & 90.9$_{\pm 0.6}\%$&  89.9$_{\pm 1.1}\%$                     
\\
& CTs    &86.7$_{\pm 1.1}\%$                      & 87.0$_{\pm 1.6}\%$                       &  85.5$_{\pm 1.5}\%$       
                          \\\hline
\multirow{2}{*}{Ours}                 & X-rays                             &\textbf{96.8}$_{\pm 0.2}\%$                  &\textbf{97.5}$_{\pm 0.2}\%$                     & \textbf{95.9}$_{\pm 0.2}\%$                      \\
& CTs                                & \textbf{93.2}$_{\pm 0.9}\%$                     &\textbf{94.1}$_{\pm 0.8}\%$                        & \textbf{92.4}$_{\pm 0.9}\%$                            \\ \hline
\end{tabular}}
\end{table}

Furthermore, we dive into the efficiency performance of the adversarial masking served as the \emph{ante-hoc} interpretation tool. The comparison of the explanation time per instance between the attention map with Grad-CAM and the adversarial masking is shown in Table \ref{time}. It illustrates that the adversarial masking compares quite superiorly against the attention map which restrictively paints the attention weights through a whole encoder. Contrarily, due to
the multifarious cross-modal masks are generated before the encoder, the explanation time per instance overhead of adversarial mask exceeds $20\%$ and $25\%$ of vision and question respectively compared with the attention map, indicating that the adversarial masking improves the interpretability for both accuracy and efficiency.

The particular comparison obviously demonstrates that the unified adversarial masking strategy motivates the learning of unified image and text representations with explanation, generating promising multifarious semantic parts rather than random masks to predict and interpret with remarkable performance and efficiency. The further prominent is that, with unified representations,  the learned vision and question masks are closely associated for improving the \emph{ante-hoc} interpretability.

\subsection{An Additional Discussion About the Performance of UnICLAM in Diagnosing Practical Heart Failures }
Due to the discrepancy  between practical medical cases and  available Medical-VQA datasets chosen from online medical libraries, we newly conduct a practical Medical-VQA dataset for heart failure diagnosing and take an 
additional discussion on the performance of UnICLAM in it.
We first sampled the images which were collected from actual inspection heart failure (HF) patients with detailed clinical examination between April 2021 and November 2022. Images from patients without HF between October and November 2022 were also collected. This anonymized database included 97 anteroposterior or lateral chest X-rays, 800 chest CTs of different layers. Left ventricular ejection fraction (LVEF) and brain natriuretic peptide (BNP) levels are using to confirm diagnosis.
For each image, we provide two related medical questions which are $Q1$:\emph{"Is the pulmonary vascular shadow thickening?"} and $Q2$:\emph{"Is the heart shadow abnormally large?"} with the help of experienced doctors. Only if the model answers two questions with \emph{yes}, the corresponding image is diagnosed as heart failure. With the pre-trained model on VQA-RAD, we select $5$ chest X-rays, $10$ chest CTs as the trainset,  and the rest is the testset.

As shown in Table \ref{pratical}, we compare UnICLAM with 3 state-of-the-art Medical-VQA models. It is apparently seen that the UnICLAM has conspicuous results, 
yielding 95.9$\%$ and 92.4$\%$ overall accuracy on the practical X-rays and CTs datasets respectively.
Specifically, there are $10.7\%$ and $11.2\%$  accuracy improvements obtained by UnICLAM for X-rays and CTs datasets compared with the MEVF+BAN \citep{mevf_san_nguyen2019overcoming}. Furthermore, compared with the MMBERT \citep{khare2021mmbert} which has constructed a generalized cross-modal fusion model, the UnICLAM with unified vision and texture representations gains $9.1\%$ and $9.5\%$  overall accuracy improvements. Besides, the UnICLAM considerably surpasses the advanced model PubMedCLIP \citep{clipeslami2021does} by $6.0\%$ and $6.9\%$ accuracy on X-rays and CTs datasets respectively (the accuracy of $Q1$ considerably increased by  $5.8\%$, $6.5\%$ and $Q2$ increased by $6.6\%$, $7.1\%$ respectively).
As the experiments indicate, our UnICLAM has the superiority of the few-shot adaption in practical Medical-VQA datasets.

As the Fig. \ref{fig:practical mask} shows, we also construct the   comparison of interpretability on the conducted practical Medical-VQA datasets.
For the CT (row $1$) and X-ray (row $3$) of chest organ with $Q2$, the learned question adversarial masks  precisely focus on  the organ \emph{heart} and symptoms \emph{abnormally, large}. Meanwhile, the adversarial vision mask learns to mask specific object parts like the heart shadow (orange mask in the center of CT, X-ray), 
while MEVF+BAN \citep{mevf_san_nguyen2019overcoming}  focuses on meaningless words like \emph{"Is"} and misleads the visual attention to the irrelevant black background. Besides, compared with the attention map of UnICLAM which focuses on the only relevant area of the heart, the adversarial masks precisely locate the outline of the heart shadow and associated entities in different colors for a convincing result. Moreover, the adversarial vision and texture masks are precisely semantically corresponding while the original masks are not.

The second column introduces the normal CT sample.
In comparison,  the proposed method predicts the right answer with the proper question keywords and concrete vision mask regions of the pulmonary vascular shadow scattered around the heart, while the MEVF+BAN \citep{mevf_san_nguyen2019overcoming} inversely ignores. Moreover, it can be obviously witnessed that the adversarial masks indicate orange pulmonary vascular shadows 
while the attention map of UnICLAM only highlights the unilateral lung and mixed with background interference. What's more, the adversarial masks of vision and question are demonstrated in a  correlative way compared with the meaningless origin masks.


The interpretability comparison results illustrate that our proposed method has strong few-shot adaption performance to generate semantically vision and texture masks which are closely related and further improve the  \emph{ante-hoc} interpretability with the extension to the practical disease diagnosis.


\section{Conclusion}
In this paper, we construct a unified and interpretable structure (UnICLAM) through contrastive representation learning with adversarial masking for Medical-VQA tasks. We first construct a unified dual-stream structure through constraining vision and texture encoders into the same space to gradually share parameters. This structure proves to be effective in promoting cross-modal representation alignments. Within this unified structure, we further design an adversarial masking data augmentation strategy employed for the unified contrastive representation learning of visual and texture encoders. The evaluation of the proposed method shows that this strategy achieves remarkable performance and efficiency in promoting unified representations and \emph{ante-hoc} interpretability. The proposed framework shows superior performance in terms of the VQA-RAD and SLAKE benchmarks than the existing state-of-the-art models. 
To explore this general framework in real-world problems, we further discuss the performance of UnICLAM in diagnosing heart failures. Experimental results obtained from heart failure data precisely indicate the strong few-shot adaption performance of the proposed model.

In future exploration, we plan to contribute a professional Medical-VQA dataset with practical medical images and annotated question-answer pairs from clinical examinations, so as to facilitate the real-world applications of Medical-VQA systems. Also, we would like to make a deep exploration into the open-ended set of the Medical-VQA model.

\section*{Declaration of competing interest}

The authors declare that they have no known competing financial interests or personal relationships that could have appeared to
influence the work reported in this paper.

\section*{Acknowledgments}
This work was supported in part by Zhejiang Provincial Natural Science Foundation of China (LDT23F02023F02). 
\bibliographystyle{model2-names.bst}\biboptions{authoryear}
\bibliography{refs.bib}

\end{document}